\UseRawInputEncoding
\documentclass[10pt,twocolumn,letterpaper]{article}

\usepackage[final]{cvpr}      

\usepackage{graphicx}
\usepackage{amsmath}
\usepackage{amssymb}
\usepackage{booktabs}

\newcommand{\keywords}[1]{\vspace{1em}\noindent\textbf{Keywords:} #1}

\usepackage[capitalize]{cleveref}
\crefname{section}{Sec.}{Secs.}
\Crefname{section}{Section}{Sections}
\Crefname{table}{Table}{Tables}
\crefname{table}{Tab.}{Tabs.}

\def\cvprPaperID{*****} 
\def\confName{CVPR}
\def\confYear{2022}

\begin{document}

\title{GFT: Gradient Focal Transformer}

\author{
\small\sffamily
\begin{tabular}{cc}
\textbf{Boris Kriuk} & \textbf{Simranjit Kaur Gill} \\
Hong Kong University of Science and Technology & University of Westminster \\
Dept. of Computer Science \& Engineering & Dept. of Computer Science \& Engineering \\
\texttt{\scriptsize bkriuk@connect.ust.hk} & \texttt{\scriptsize w1931485@my.westminster.ac.uk} \\[2ex]
\textbf{Shoaib Aslam} & \textbf{Amir Fakhrutdinov} \\
University of Engineering \& Technology Lahore & Shanghai Jiao Tong University \\
Dept. of Mechanical, Mechatronics \& Control Eng. & Dept. of Mechanical Engineering \\
\texttt{\scriptsize shoaib.aslam@uet.edu.pk} & \texttt{\scriptsize amir\_fakhrutdinov@sjtu.edu.cn}
\end{tabular}
}
\maketitle

\begin{abstract}
   Fine-Grained Image Classification (FGIC) remains a complex task in computer vision, as it requires models to distinguish between categories with subtle localized visual differences. Well-studied CNN-based models, while strong in local feature extraction, often fail to capture the global context required for fine-grained recognition, while more recent ViT-backboned models address FGIC with attention-driven mechanisms but lack the ability to adaptively focus on truly discriminative regions. TransFG and other ViT-based extensions introduced part-aware token selection to enhance attention localization, yet they still struggle with computational efficiency, attention region selection flexibility, and detail-focus narrative in complex environments. This paper introduces GFT (Gradient Focal Transformer), a new ViT-derived framework created for FGIC tasks. GFT integrates the Gradient Attention Learning Alignment (GALA) mechanism to dynamically prioritize class-discriminative features by analyzing attention gradient flow. Coupled with a Progressive Patch Selection (PPS) strategy, the model progressively filters out less informative regions, reducing computational overhead while enhancing sensitivity to fine details. GFT achieves SOTA accuracy on FGVC Aircraft, Food-101, and COCO datasets with 93M parameters, outperforming ViT-based advanced FGIC models in efficiency. By bridging global context and localized detail extraction, GFT sets a new benchmark in fine-grained recognition, offering interpretable solutions for real-world deployment scenarios. 
\end{abstract}


\section{Introduction}
\label{sec:intro}
Deep learning has advanced significantly in the past decade, specifically in the domain of image classification. Fine-grained image classification (FGIC) aims to identify subtle distinctions among similar classes, such as different species of birds or various car models. Such tasks are considered to be fine-grained because the model is required to distinguish between high-complexity differences in visual appearance and patterns, making it more challenging than regular image classification. The advancements in deep learning architectures, coupled with the growth of annotated datasets, have historically propelled FGIC to new heights, enabling continuous refinement of SOTA classification systems.

The origins of FGIC can be traced back to the early 2000s when traditional feature engineering approaches dominated the field. Methods such as Scale-Invariant Feature Transform (SIFT) \cite{790410}, Histogram of Oriented Gradients (HOG) \cite{1467360}, and texton-based representations \cite{790379} were combined with machine learning classifiers like SVMs to capture subtle inter-class differences. However, such handcrafted features struggled with the inherent complexity of fine-grained tasks, where discriminative cues often reside in localized regions and are easily obscured by background clutter, pose variations, or lighting changes.

The new era of deep learning started with the rise of convolutional neural networks (CNN) in the research field. The integration of convolutional layers with sliding kernel technology allowed the models to automatically capture features. Pioneering architectures like AlexNet \cite{simonyan2014very} demonstrated the power of deep hierarchical representations, while VGGNet \cite{he2016deep} and ResNet \cite{7410527} advanced accuracy through increased depth and residual connections later. However, CNN-backboned models struggled to prioritize discriminative regions without explicit guidance, prompting methods like Part-Cracked CNNs \cite{xie2017aggregated} and bilinear pooling (B-CNN) \cite{dosovitskiy2020image} to fuse multi-scale features or model part-attribute relationships. However, such approaches often require costly annotations.

The field shifted toward efficiency and scalability with ResNeXt \cite{touvron2021training} which introduced grouped convolutions for parameter reuse, and Vision Transformers \cite{9879745}, which replaced convolutions with patch-based self-attention for global-local dependency modeling. ViT’s reliance on large datasets was mitigated by DeiT \cite{Brock2021HighPerformanceLI}, which leveraged distillation and augmentation for data efficiency. 

Subsequent innovations like ConvNeXt \cite{liu2021swin} modernized CNNs using transformer-inspired techniques such as layer normalization, while NFNet \cite{NEURIPS2021_20568692} eliminated batch normalization for stability on smaller datasets. Hybrid architectures, such as Swin Transformer \cite{liu2021swin} and CoAtNet \cite{NEURIPS2021_20568692}, combined shifted-window attention with convolutional priors to enhance multi-scale reasoning. More recent models like MaxViT \cite{tu2022maxvit} and BiT \cite{10.1007/978-3-030-58558-7_29} further optimized scalability, with BiT demonstrating that pre-trained ResNet variations can rival ViTs when scaled appropriately.

Despite the latest advancements, critical challenges remain unresolved. Many architectures, particularly vision transformers and smaller-scale CNN models, remain heavily dependent on high-quality labeled datasets, restricting their utility in domains like medical imaging where annotations are  still scarce. Furthermore, conventional CNNs and static attention mechanisms often lack dynamic mechanisms to prioritize discriminative features, leading to suboptimal balances between local detail extraction and global contextual understanding–a gap evident in fine-grained tasks where subtle inter-class differences are overshadowed by irrelevant patterns. Hierarchical vision transformers and hybrid architectures, despite their performance, frequently incur prohibitive memory and processing costs, limiting real-world deployments. Additionally, rigid token selection strategies in standard vision transformers hinder generalization under real-world variations such as occlusion or domain shifts, while parameter-efficient CNNs \cite {mostafa2019parameter} and neural architecture search-derived models \cite {ren2021comprehensive} struggle to preserve pixel-level precision in tasks requiring granular detail.

Moreover, ViT-backboned models face additional challenges that affect their efficiency and performance. In scenarios with highly variable object scales or complex spatial relationships ViT-derived models' reliance on fixed tokenization strategies results in inefficient feature extraction \cite {toraman2023impact}. Such behavior becomes a big drawback for FGIC tasks, where precise boundaries matter. Hybrid models try to fix the mentioned problem by combining convolutional layers with transformers, which leads to increase in accuracy; however still reintroduces local biases and limits global receptive field advantage of transformers \cite{bhattamishra2020ability}. 

The quadratic complexity of self-attention remains a major bottleneck, particularly for high-resolution imagery, where computational overhead restricts practical scalability. The large memory footprint required to store attention maps and intermediate activations further restraints batch size choice during training, slowing convergence process \cite {mccandlish2018empirical}. Current limitations collectively underscore the need for models that adaptively focus on task-critical regions, reduce computational redundancy, and maintain robustness across diverse data regimes. 

We propose GFT, a new ViT-backboned framework designed to address existing gaps through two core innovations: 
\begin{enumerate}
    \item hierarchical attention module, which we called \textit{GALA (Gradient Attention Learning Alignment)}, that combines multi-head attention with gradient-based importance scoring and window-based region focus to dynamically prioritize features across local and global scales,
    \item multi-scale processing that employs dynamic patch selection to extract features from general to fine-grained details, needed for FGIC task success.
\end{enumerate}

GFT aims to tackle real-world fine-grained imagery challenges through an efficient transformer architecture that dynamically selects critical features while maintaining global context.

\section{Related Works} \label{sec:literature}
This section provides a concise summary of the endeavors undertaken to tackle the challenge of image classification.

\subsection{Image Classification with Convolutional Neural Networks}
The evolution of image classification began with foundational works like \cite{fukushima1980neocognitron}, who introduced the Neocognitron, a model inspired by the visual cortex, and \cite{lecun1989backpropagation, lecun1998gradient}, who developed LeNet-5, one of the first practical CNNs for handwritten digit recognition. These early works laid the groundwork for modern deep learning. The field saw a revolution with AlexNet \cite{krizhevsky2012imagenet}, which demonstrated the power of deep CNNs by winning the ImageNet competition, popularizing GPUs for training. This was followed by \cite{simonyan2014very}, who introduced VGGNet, emphasizing depth, and \cite{szegedy2015going}, who proposed GoogLeNet with its Inception modules for multi-scale feature extraction. Further, ResNet \cite{he2016deep} advanced by introducing residual connections to enable training of very deep networks. Architectures like  DenseNet \cite{huang2017densely}'s and ResNeXt \cite{xie2017aggregated} improved feature reuse and scalability. Efficient CNNs emerged with MobileNet \cite{howard2017mobilenets} and MobileNetV2 \cite{sandler2018mobilenetv2}, which optimized for mobile devices using depthwise separable convolutions. \cite{zhang2018shufflenet} introduced ShuffleNet, which used channel shuffling to reduce computational cost, while \cite{ma2018shufflenet} proposed ShuffleNet V2, offering practical guidelines for efficient CNN design. \cite{tan2019efficientnet} rethought model scaling with EfficientNet, achieving state-of-the-art performance.

Attention mechanisms were integrated into CNNs by \cite{wang2018non} with Non-Local Networks and with SENet \cite{hu2018squeeze}, which recalibrated channel-wise features. \cite{woo2018cbam} further refined attention with CBAM (Convolutional Block Attention Module), combining spatial and channel attention, and \cite{bello2019attention} introduced attention-augmented convolutional networks. While these advancements significantly improved performance, they also highlighted challenges such as computational cost, scalability, and the need for large labeled datasets.

\subsection{Transformers in Vision}
The introduction of transformers in vision began with \cite{vaswani2017attention}, who proposed the transformer architecture for natural language processing (NLP), inspiring its adaptation to computer vision. \cite{dosovitskiy2020image} introduced Vision Transformers (ViTs), demonstrating that the transformers could achieve image classification performance by treating images as a sequence of patches. The work \cite{touvron2021training} improved the training efficiency of ViTs with DeiT, using knowledge distillation to reduce the data requirements. \cite{chen2021generative} explored generative pretraining with Image GPT, extending transformers to unsupervised image modeling. \cite{liu2021swin} proposed Swin Transformers, which introduced shifted windows to enable hierarchical feature learning and scalability to high-resolution images. \cite{yuan2021tokens} introduced T2T-ViT, which progressively aggregated tokens to capture local structure, while \cite{wu2021cvt} combined the strengths of CNNs and transformers with CvT. While these works demonstrated the potential of transformers in vision, they often required large-scale datasets and computational resources, limiting their accessibility. Additionally, the interpretability of transformer-based models remained a challenge, as their attention mechanisms were less intuitive compared to CNNs.

The work \cite{wang2021pyramid}, proposed PVT, a pyramid vision transformer for dense prediction tasks. Scaling transformers was explored by \cite{zhai2021scaling}, who studied the impact of scaling ViTs in terms of depth, width, and resolution. \cite{touvron2021going} introduced LeViT, a hybrid CNN-Transformer architecture optimized for efficiency, and \cite{mehta2022mobilevit} proposed MobileViT, combining the strengths of CNNs and transformers for mobile devices. Self-supervised learning with transformers was advanced by \cite{caron2021emerging}, who introduced DINO (Self-Distillation with No Labels), and \cite{he2022masked}, who proposed MAE, a masked autoencoder framework for scalable vision learning. \cite{bao2021beit} explored BERT-style pretraining for image transformers with BEiT, while \cite{chen2021empirical} studied self-supervised training of ViTs. These works addressed data efficiency and scalability, but they often relied on complex pretraining strategies, which could be computationally expensive. Moreover, the performance gains were sometimes marginal compared to the increased complexity of the models.

Multimodal transformers were introduced by \cite{radford2021learning} with CLIP (Contrastive Language-Image Pretraining) , which aligned text and image embeddings, and \cite{alayrac2022flamingo}, who proposed Flamingo for few-shot visual-language learning. \cite{li2021align} introduced ALIGN, which aligned vision and language representations using contrastive learning, and \cite{yu2022coca} proposed CoCa, a contrastive captioner for image-text foundation models. Hybrid architectures such as \cite{yuan2021tokens}, \cite{wu2021cvt}, and \cite{chen2021crossvit} combined CNNs and transformers to leverage their complementary strengths. \cite{tolstikhin2021mlp} introduced MLP-Mixer, which replaced attention with MLPs, and \cite{touvron2021resmlp} proposed ResMLP, a simpler alternative to transformers. \cite{tu2022maxvit} introduced MaxViT, which combined multi-axis attention with convolutional layers for improved efficiency and performance. Multimodal transformers demonstrated impressive capabilities in unifying vision and language, but they often required massive datasets and computational resources. Additionally, their performance in low-resource settings remained suboptimal, highlighting the need for more efficient and accessible approaches. MaxViT and similar hybrid models showed promise in balancing efficiency and performance, but their scalability to larger datasets and tasks remained to be fully explored.

\section{Methodology}
\subsection{Architecture Overview}

At its core, GFT builds upon the established Vision Transformer framework while introducing new vision that enables adaptive feature selection based on gradient information flows.

The GFT architecture comprises three main components: (1) a base transformer encoder derived from ViT, (2) hierarchical attention blocks (GALA) that compute gradient-based importance, and (3) a progressive patch selection mechanism that iteratively refines the visual representation by focusing on the most informative regions of the input image.

The model processes an input image of size 224×224 pixels through a standard patch embedding layer, dividing the image into non-overlapping patches of size 16×16, resulting in a sequence of 196 patch tokens. A learnable class token is prepended to this sequence following the standard ViT approach. The resulting tokens are processed through the initial 8 transformer blocks to establish a foundational representation of the visual content.

After the initial processing, the model enters its distinctive hierarchical refinement phase. This phase consists of three GALA blocks (Fig 1.)  that progressively analyze and filter the patch tokens. Each block applies a specialized attention mechanism that not only updates the token representations but also computes gradient-based importance scores that identify patches containing the most valuable visual information.

The importance scores derived from the attention gradients serve as the basis for the progressive patch selection strategy. In each hierarchical stage, the model selects a decreasing subset of the patches (75\%, 50\%, and 25\% of the original patches in the three stages, respectively) based on their computed importance. The progressive reduction allows the model to focus computational resources on the most informative regions of the image while discarding redundant or uninformative patches.

\begin{figure*}
    \centering
    \includegraphics[height=10cm]{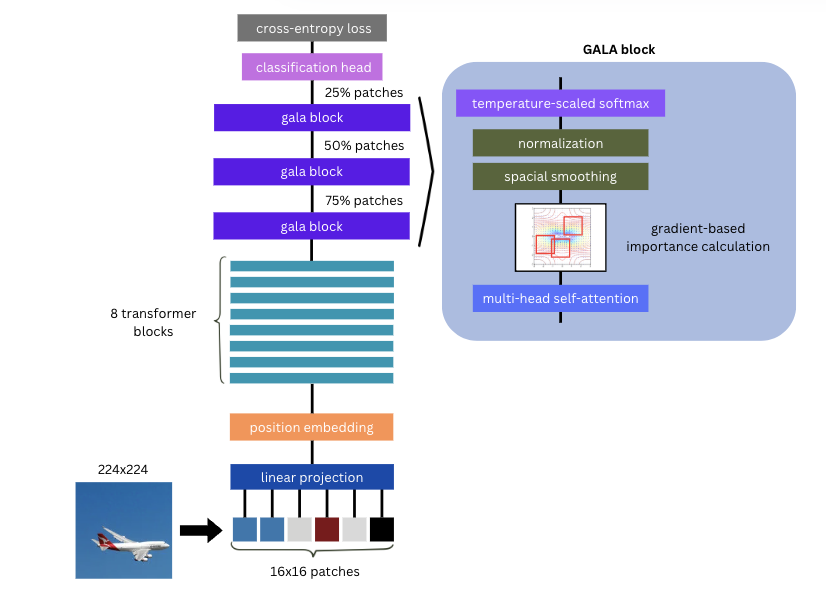}
    \caption{GFT Architecture Overview.}
    \label{fig1}
\end{figure*}

The final classification is performed using the class token, which aggregates information from the refined patch tokens through the attention mechanism. The token passes through a final classification head consisting of a linear layer that maps to the target number of classes.

Despite this relatively small increase in model size, the hierarchical design enables GFT to achieve more efficient processing by focusing on informative patches, reducing computational cost during inference while improving classification accuracy depth.

Our architectural design maintains full compatibility with existing ViT pre-trained weights, allowing the model to leverage transfer learning from models trained on large-scale datasets like ImageNet. During training, the model benefits from both the pre-trained weights of the base transformer blocks and the adaptive capability of the newly introduced attention mechanism, enabling efficient fine-tuning on downstream tasks.

We expand on GALA and progressive patch selection in sections 2.2 and 2.3 respectively.

\subsection{GALA: Gradient Attention Learning Alignment}

Traditional approaches with ViT-backbone models rely predominantly on absolute attention values to identify salient regions. This methodology operates under the implicit assumption that the magnitude of attention directly correlates with feature importance. However, this idea is clearly oversimplified for fine-grained image classification and does not reflect the full complexity of visual context. We state that it is not the absolute attention values itself but the rate of change in attention value distribution across spatial locations that encodes the most discriminative information for fine-grained tasks.

\begin{figure}
\begin{center}
\includegraphics[height=4cm]{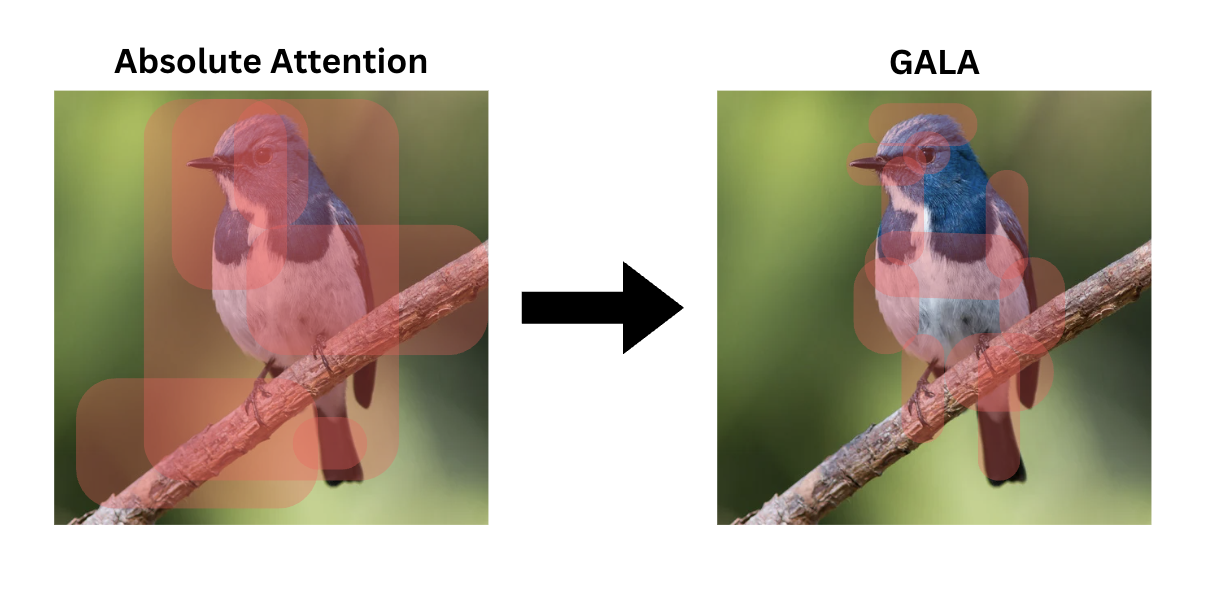}
\end{center}
\caption 
{ \label{fig1}
Absolute Attention vs GALA.} 
\end{figure} 

The Gradient Attention Learning Alignment (GALA) mechanism is implemented in FGT and formalizes a new perspective by computing gradients across the attention landscape to identify regions of high variation—precisely where the most class-discriminative features typically reside, as shown in Fig. 2. Given an attention score matrix from transformer heads, we first compute the mean attention across the target dimension to obtain a more stable representation of attentional focus and reduce the noise. Such aggregation prevents individual outlier values from disproportionately influencing the importance assessment (1).

\begin{equation}
\bar{\mathbf{A}}{b,h,i} = \frac{1}{N}\sum{j=1}^{N}\mathbf{A}_{b,h,i,j}
\end{equation}

The core of GALA lies in calculating the spatial gradient of this attention distribution. Rather than relying on simple first-order differences, we implement a second-order central difference approximation for enhanced stability. The mathematical formulation captures not just the directional change but also the rate of change in attentional focus across the image patches, revealing boundaries between semantic regions where discriminative features often reside (2).

\begin{equation}
\nabla\bar{\mathbf{A}}_{b,h,i} = 
\begin{cases}
\bar{\mathbf{A}}_{b,h,i+1} - \bar{\mathbf{A}}_{b,h,i}, & \text{if } i = 0 \\
\frac{\bar{\mathbf{A}}_{b,h,i+1} - \bar{\mathbf{A}}_{b,h,i-1}}{2}, & \text{if } 0 < i < N-1 \\
\bar{\mathbf{A}}_{b,h,i} - \bar{\mathbf{A}}_{b,h,i-1}, & \text{if } i = N-1
\end{cases}
\end{equation}

The absolute gradient magnitude is then aggregated across attention heads to produce an importance score for each patch. The aggregation acknowledges that different attention heads may specialize in different aspects of visual recognition, and their collective gradient behavior provides a more comprehensive view of feature importance. To account for spatial continuity in visual features, we apply a 1D convolutional smoothing operation with a learned kernel. Such smoothing is essential because adjacent patches often contain related information, and abrupt changes in importance scores may reflect noise rather than meaningful feature boundaries. The learned nature of the smoothing filter allows the model to adapt its spatial integration behavior to the specific characteristics of fine-grained classification.

We introduce GALA with the incorporation of temporal consistency through an exponential moving average (EMA) mechanism that stabilizes importance scores during training. Our temporal smoothing implementation recognizes that importance patterns should evolve gradually across training iterations, preventing excessive oscillation that might otherwise hinder convergence. The normalization terms ensure statistical stability, preventing the domination of a few extreme values.

The final importance distribution is obtained through temperature-scaled softmax normalization. The temperature parameter controls the sharpness of the distribution, allowing for more selective focus on the most informative regions. A lower temperature value increases the contrast between high and low importance regions, effectively amplifying the model's decisiveness in patch selection (3).

\begin{equation}
\mathcal{P}{b,i} = \frac{\exp(\mathcal{I}^{\text{EMA}}{b,i}/\tau)}{\sum_{j=1}^{N}\exp(\mathcal{I}^{\text{EMA}}_{b,j}/\tau)}
\end{equation}

The new formulation represents a fundamental shift from existing approaches—instead of concentrating on where attention is highest, GALA focuses on where attention is changing most rapidly. Our ideas align with the intuition that discriminative features in fine-grained tasks often exist at the boundaries between different semantic regions (e.g., wing-to-fuselage transitions in aircraft, or beak-to-head boundaries in bird species). By explicitly modeling these transitions through gradient-based importance, GALA overcomes the limitations of absolute attention mechanisms that may disproportionately focus on dominant but non-discriminative features.

\subsection{PPS: Progressive Patch Selection}

The Progressive Patch Selection (PPS) mechanism addresses critical limitations in existing ViT-backboned fine-grained approaches by implementing a different paradigm for computational efficiency in transformer-based architectures. Current state-of-the-art methods, such as TransFG, have demonstrated the potential of transformer architectures for fine-grained classification by incorporating attention-based feature selection. TransFG utilizes a single-stage selection approach that identifies informative patches through a jigsaw-inspired attention mechanism, which selects a fixed number of patches for subsequent processing. While effective, this single-stage selection strategy fails to capture the hierarchical nature of discriminative features that characterize fine-grained recognition tasks.

In contrast to TransFG's one-time patch selection, PPS implements a multi-stage refinement process that mirrors the progressive examination strategies employed by human experts when differentiating similar object categories. When distinguishing between automobile models, for example, humans typically first identify the general vehicle type before progressively narrowing their focus to specific discriminative details such as headlight shape, grille design, or badge placement. Such natural coarse-to-fine attention pattern is algorithmically encoded in PPS through its staged selection approach, which systematically reduces the patch count while increasing the concentration of informative content.

Other approaches like RAM-ViT and DCT have attempted to address computational efficiency in vision transformers through frequency-domain transformations or random patch sampling. However, these methods lack the adaptive, content-aware selection mechanism that makes PPS particularly suited for fine-grained tasks. RAM-ViT, for instance, applies a uniform compression strategy across all images regardless of their content, potentially discarding discriminative details essential for fine-grained classification. DCT-based approaches reduce dimensionality but maintain the same number of tokens throughout the network, limiting their computational gains. PPS, by contrast, adapts its selection process to the specific content of each image, ensuring that computational resources are allocated precisely where they provide maximum discriminative value (4).

\begin{align}
S_i = {p_{j} | \text{rank}(I(p_{j})) \leq k_i \cdot N, j \in {1,2,...,N}} \tag{4}
\end{align}

The progressive focus on increasingly detailed visual elements not only aligns with cognitive models of visual expertise but also yields substantial computational benefits (5). By retaining only 75\%, then 50\%, and finally 25\% of patches at successive stages, as shown in Fig. 2, PPS dramatically reduces the quadratic computational complexity that typically bottlenecks transformer architectures. We find the reduction particularly useful in deeper network layers, where self-attention operations across numerous tokens constitute the majority of computational overhead. Traditional vision transformers must maintain all patches throughout the network, resulting in computational requirements that scale quadratically with the number of patches. PPS disrupts this unfavorable scaling by progressively pruning less informative patches, focusing computational resources exclusively on regions with high discriminative potential (6).

\begin{align}
I(p_j) = \frac{1}{H} \sum_{h=1}^{H} |\nabla_{p_j} A_h|_F \tag{5}
\end{align}

\begin{align}
C_{PPS} = C_{base} \cdot \left(1 - \sum_{i=1}^{L} \alpha_i \cdot (1 - k_i)\right) \tag{6}
\end{align}

\begin{figure}
\begin{center}
\includegraphics[height=1.8cm]{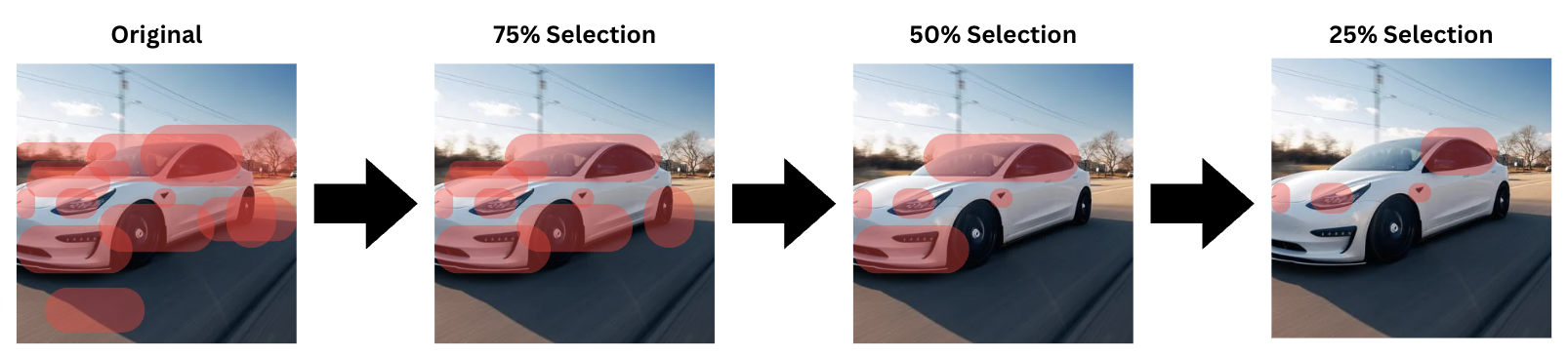}
\end{center}
\caption 
{ \label{fig3}
Progressive Patch Selection in GFT.} 
\end{figure} 

The effectiveness of PPS can be attributed to its alignment with the inherent structure of fine-grained classification tasks, where discriminative information is often concentrated in a small subset of the visual field. By iteratively refining its focus to these informative regions, PPS reduces the model's tendency to overfit to background elements or shared features that do not contribute to class discrimination. The targeted allocation of computational resources allows the network to develop more specialized representations of class-specific details, enhancing the ability to distinguish between similar categories.

\section{Experiments}

We perform experiments with popular SOTA base models to measure the theoretical potential of different approaches in FGIC. We strive to diversify the experiment pool, choosing datasets with different contextual data representation and image high information saturation levels. For example, we avoid some popular FGIC datasets due to lower levels of information saturation: smaller image sizes, resolution or number of classes. Datasets such as FGVC Aircraft \cite{maji2013fine}, Food-101 \cite{bossard2014food}, COCO \cite{lin2014microsoft} have been chosen for final evaluation.

The FGVC Aircraft dataset presents a significant challenge in fine-grained visual classification with 10,000 images spanning 100 aircraft model variants. This dataset demands discriminative capability as it requires models to distinguish between visually similar aircraft categories based on subtle variations in wing configuration, tail design, engine placement, and fuselage structure. With its high-resolution images captured from multiple viewpoints and under varying conditions, FGVC Aircraft serves as an ideal testbed for evaluating a model's ability to identify fine-grained visual details that are critical for accurate classification in specialized domains. The results are in Table 1.

The ViT Base establishes a solid foundation with 65.86\% accuracy, while TransFG Base achieves a substantial improvement to 76.45\% by incorporating part-aware token learning mechanisms that enhance the model's ability to focus on discriminative regions. TransFG's success comes from its part-aware training strategy that encourages attention to informative local regions through token drop mechanisms, allowing it to better capture the differences between aircraft variants. The GFT Base model further refines results, achieving marginally better performance (76.51\% accuracy) while using fewer parameters than TransFG base model (93M vs. 101M). FGT's superior performance demonstrates increased generalization level and stronger theoretical feature-learning capabilities. GFT manages to improve the straightforward yet effective focus model of TransFG by choosing the focus regions that contribute to the class differentiation only.

Among the CNN-based models, DenseNet169 stands out with the highest accuracy of 76.33\% and an F1-score of 0.7659, demonstrating its ability to effectively propagate features and gradients over dense connections. The model shows balance between accuracy and computational efficiency, as it matches the performance of larger transformer models while having only 14.3 million parameters.  On the other hand, BiT, despite its large-scale pretraining with 382 million parameters, achieves only 69.01\% accuracy, reflecting challenges in transferring generic visual priors to highly specialized fine-grained tasks. ResNet18, the smallest CNN with 11 million parameters, lags significantly at 56.17\% accuracy, highlighting the limitations of shallow architectures for fine-grained discrimination. Overall, while CNNs remain viable for structured image recognition tasks, they do not offer a scalable pathway to generalization similar to GFT.

\begin{table}[ht!]
\centering
\caption{FGVC Aircraft Model Performance Comparison.}
\scriptsize
\begin{tabular}{l rrrr l}
\toprule
Model & Acc & Prec & Rec & F1 & Type \\
\midrule
ViT-B & 65.9\% & 0.662 & 0.659 & 0.657 & ViT-86M \\
DeiT-B & 73.2\% & 0.739 & 0.732 & 0.733 & ViT-86M \\
TransFG & 76.5\% & 0.771 & 0.765 & 0.764 & ViT-101M \\
ResNet18 & 56.2\% & 0.571 & 0.562 & 0.559 & CNN-11M \\
BiT & 69.0\% & 0.697 & 0.690 & 0.689 & CNN-382M \\
DenseNet & 76.3\% & 0.778 & 0.763 & 0.766 & CNN-14M \\
\textbf{GFT-B} & \textbf{76.5\%} & \textbf{0.774} & \textbf{0.764} & \textbf{0.765} & \textbf{ViT-93M} \\
\bottomrule
\end{tabular}
\label{tab:model_comparison}
\end{table}

\begin{table}[ht!]
\centering
\caption{Food 101 Model Performance Comparison}
\scriptsize
\begin{tabular}{l rrrr l}
\toprule
Model & Acc & Prec & Rec & F1 & Type \\
\midrule
ViT-B & 74.86\% & 0.749 & 0.749 & 0.748 & ViT-86M \\
DeiT-B & 78.61\% & 0.787 & 0.786 & 0.786 & ViT-86M \\
TransFG & 79.77\% & 0.798 & 0.798 & 0.797 & ViT-101M \\
ResNet18 & 73.76\% & 0.738 & 0.738 & 0.737 & CNN-11M \\
BiT & 80.38\% & 0.805 & 0.804 & 0.804 & CNN-382M \\
DenseNet & 80.82\% & 0.809 & 0.808 & 0.809 & CNN-14M \\
\textbf{GFT-B} & \textbf{80.83\%} & \textbf{0.809} & \textbf{0.808} & \textbf{0.809} & \textbf{ViT-93M} \\
\bottomrule
\end{tabular}
\label{tab:food101_comparison}
\end{table}

The Food-101 dataset was chosen as it is a widely recognized benchmark for advancing visual recognition. It introduces a complex challenge in fine-grained image classification with 101,000 images categorized into 101 distinct food classes. Each class contains 1,000 images collected from the net, exhibiting high intra-class variation and noise due to differences in presentation, ingredients, lighting, and image background. As the dataset was designed to reflect real-world complexity, a variety of features make it suitable for evaluating the robustness and accuracy of visual recognition models. Results based on the Food-101 dataset can be seen in Table 2.

The ViT Base model showed a baseline of 74.86\% accuracy and F1-score of 0.7477, being the lightest model among the ViT family with 86M parameters. Despite a strong self-attention mechanism, ViT does not perform well when there is high intra-class variation in the data. TransFG Base improves performance with an accuracy of 79.77\% and F1-score of 0.7969, implementing attention to discriminative food features. The GFT Base model masters prediction quality even further to attain 80.83\% accuracy while keeping reasonable parameter size (GFT'S 93M vs. TransFG's 101M). GFT's enhanced ability to effectively incorporate feature learning while filtering out redundant or misleading details is crucial for classification.

CNN-based architectures displayed strong feature extraction capabilities, particularly with larger models. ResNet18 reached 73.76\% accuracy, closely aligning with ViT, however having only 11M parameters. Meanwhile, BiT (Big Transfer) achieves 80.38\% accuracy, benefiting from large-scale pretraining and transfer learning techniques. This model has the largest number of parameters and the lowest F1/size ratio (0.2103). Notably, DenseNet169 achieves the highest accuracy among CNN-based models with 83.82\%, demonstrating its efficiency in feature propagation, while maintaining solid efficiency with only 14.3M parameters. In contrast, CNNs like BiT achieve limited gains despite being scaled up, showing the need of advanced feature learning techniques.

\begin{table}[ht!]
\centering
\caption{COCO Model Performance Comparison}
\scriptsize
\begin{tabular}{l rrrr l}
\toprule
Model & Acc & Prec & Rec & F1 & Type \\
\midrule
ViT-B & 60.54\% & 0.608 & 0.605 & 0.591 & ViT-86M \\
DeiT-B & 65.07\% & 0.646 & 0.651 & 0.639 & ViT-86M \\
TransFG & 65.23\% & 0.643 & 0.652 & 0.641 & ViT-101M \\
ResNet18 & 62.52\% & 0.614 & 0.625 & 0.614 & CNN-11M \\
BiT & 64.72\% & 0.648 & 0.647 & 0.633 & CNN-382M \\
DenseNet & 64.42\% & 0.640 & 0.644 & 0.634 & CNN-14M \\
\textbf{GFT-B} & \textbf{65.83\%} & \textbf{0.650} & \textbf{0.658} & \textbf{0.645} & \textbf{ViT-93M} \\
\bottomrule
\end{tabular}
\label{tab:coco_comparison}
\end{table}

COCO represents a more challenging benchmark with 328,000 images across 91 common object categories in complex scenes. Unlike Food-101's focused domain, COCO tests model capabilities on diverse everyday objects in varied contexts, requiring robust scene understanding and object recognition.

Table 3 demonstrates consistent performance patterns across architectures. The ViT Base model achieved 60.54\% accuracy with an F1-score of 0.591, showing limitations with complex scenes. DeiT Base improved upon this foundation with 65.07\% accuracy and 0.639 F1-score through its distillation approach. The TransFG Base model performs slightly better with 65.23\% accuracy by focusing on discriminative features.

Looking at CNN-based models, DenseNet169 leads with 64.42\% accuracy and 0.634 F1-score, once again validating its efficient feature reuse strategy, even in less structured scenes. BiT, although effective with 64.72\% accuracy, is parameter-heavy (382 million), resulting in a lower parameter-efficiency ratio. ResNet18, while lightweight (11 million parameters), achieves only 62.52\% accuracy, suggesting that shallow CNNs may struggle with contextual reasoning in the COCO dataset. Such results demonstrate that balancing accuracy and scalability in complex visual tasks requires adaptive attention mechanisms as implemented in GFT and efficient propagation as exemplified by DenseNet.

\begin{figure}
    \begin{center}
        \includegraphics[width=\columnwidth]{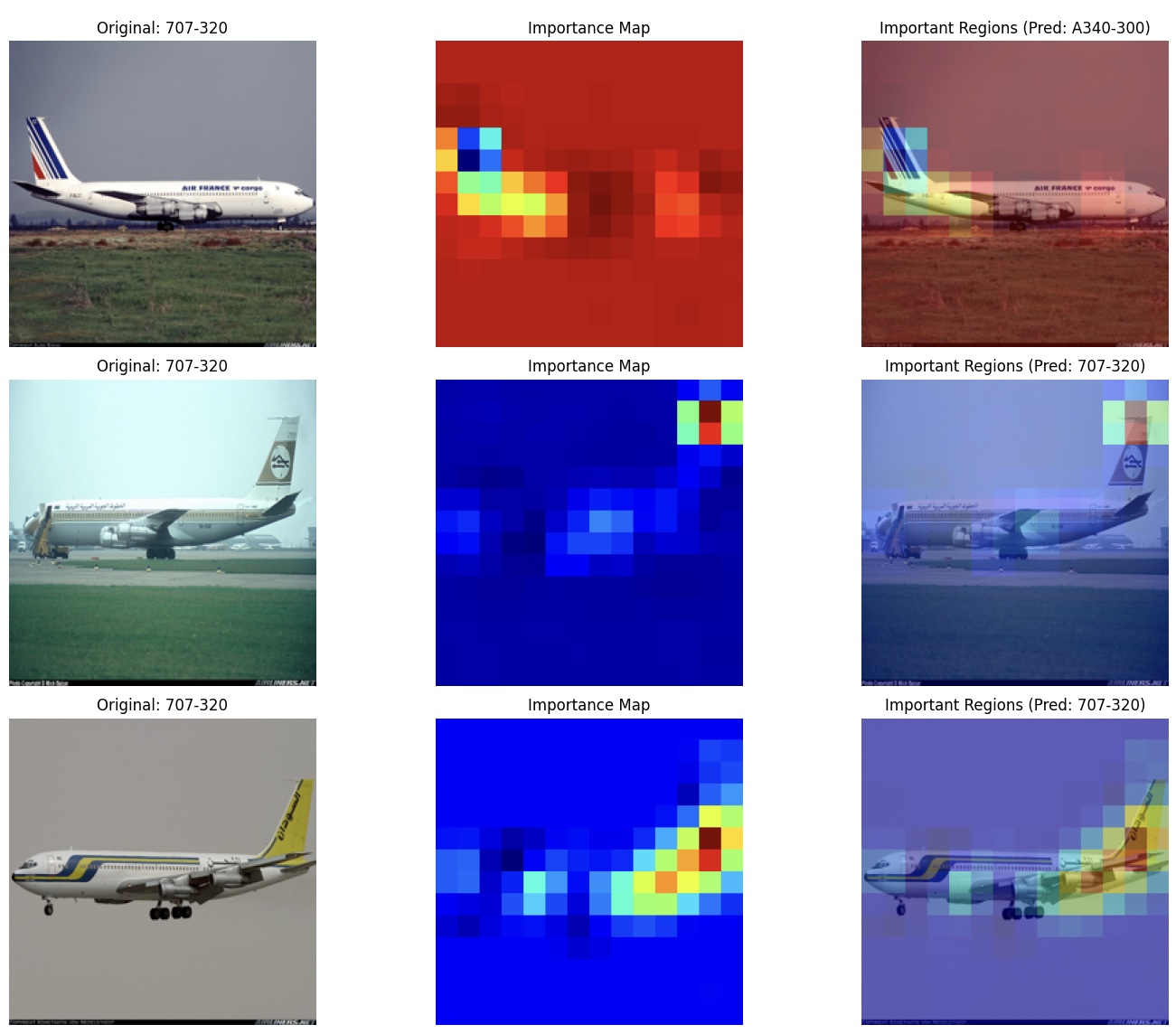}
    \end{center}
    \caption {GFT Importance Regions in FGVC Aircraft Dataset.}
    \label{fig4}
\end{figure} 

Figure 4 presents a visual interpretability technique analysis of GFT trained on FGVC Aircraft Dataset, highlighting the original aircraft image, an importance map showing the model's attention distribution, and a composite image with the regions most influential in classification decisions.

GFT adaptively transitions its focus between distinctive aircraft components—such as nose profile, wing-body junction, engine placement, and tail configuration—rather than relying on a single diagnostic feature. In the Boeing 707-320 examples, observations reveal significant shifts in the model’s attention when viewed from different angles: frontal views prominently feature the wing base, forward fuselage, and nose regions; side-profile views distinctly highlight the tail section, engines, and landing gear assemblies, indicating their importance for accurate classification. The first row illustrates an interesting misclassification case where the model predicts an A340-300 instead of the actual 707-320 due to the aircraft’s 2-engine composition, with attention concentrated at the nose and wing areas, suggesting these regions contain visually ambiguous features between two aircraft types.

The progressive feature selection approach enables GFT to achieve SOTA accuracy in comparison to baseline models while providing highly interpretable visual evidence of the feature focus selection, addressing the fine-grained recognition needed for fundamental differentiation between subtle characteristics.

\begin{figure}
    \begin{center}
        \includegraphics[width=\columnwidth]{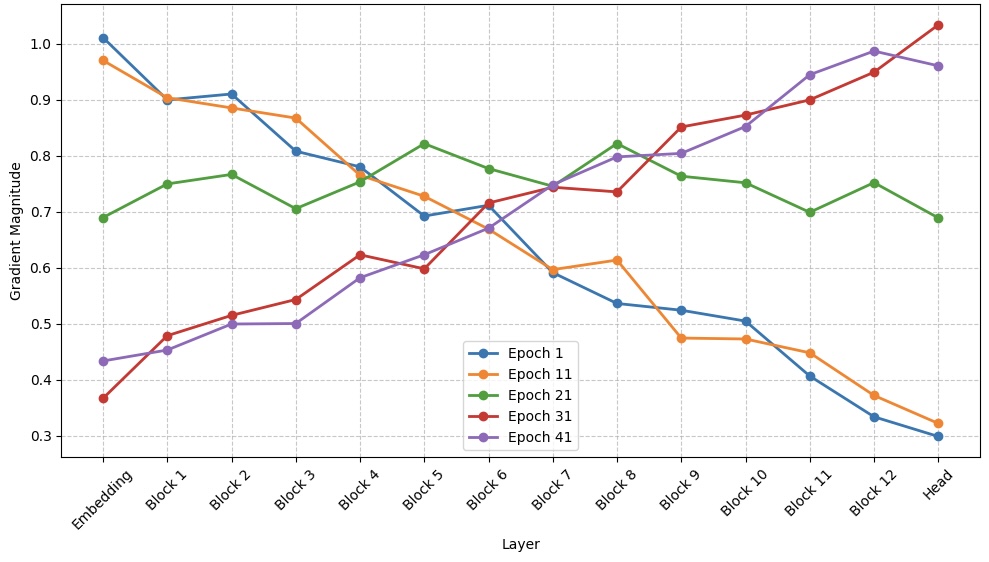}
    \end{center}
    \caption {Gradient Flow across GFT Layers.}
    \label{fig5}
\end{figure} 

The gradient flow visualization in Fig. 5 shows fundamental training dynamics, demonstrating a systematic evolution of feature importance across network depths and training epochs. Early training (epoch 1) exhibits front-loaded gradients concentrated in embedding and initial blocks, indicating the model's initial focus on basic feature extraction, with gradients diminishing in deeper layers. As training progresses through epochs 11 and 21, gradient redistribution occurs, flattening across middle blocks, suggesting a transition toward holistic feature processing. Most significantly, late-stage training (epochs 31 and 41) signifies an inverted gradient pattern compared to early epochs, with magnitudes progressively increasing through deeper layers and peaking at the classification head, precisely where the hierarchical attention mechanisms operate. 

The systematic gradient redistribution empirically validates the GFT architecture's progressive learning strategy. The network initially establishes foundational representations before shifting computational focus toward fine-grained discriminative regions via gradient-based attention. The equilibrium observed around epoch 21, where gradients stabilize uniformly across middle layers, marks a pivotal transition between general feature extraction and specialized feature refinement phases. The transition is essential, explaining the model’s capacity to adaptively attend to different structural aircraft components depending on viewpoint and orientation. 

Experiment set-up shows that GFT excels through an integrated adaptive attention selection mechanism GALA that focuses precisely on class-differentiating regions while avoiding irrelevant features. The progressive patch selection approach enables effective transition from low to high-level feature extraction during the learning process. With fewer parameters than SOTA ViT-backboned competitors like TransFG and SOTA CNN-backboned competitors like BiT, GFT delivers comparable or superior performance, better scalability than CNNs, and provides interpretable attention gradient maps valuable for regulated applications. These strengths make GFT a promising and highly practical solution for fine-grained image classification across multiple domains.

\section{Conclusion}
In this study, we introduced the Gradient Focal Transformer (GFT), a new Vision Transformer (ViT)-based framework designed to address key challenges in Fine Grained Classification (FGIC). By integrating the Gradient Attention Learning Alignment (GALA) mechanism, GFT dynamically prioritizes class-discriminative features through gradient-guided refinement, ensuring adaptive focus on the most informative regions. Additionally, the Progressive Patch Selection (PPS) approach improves computational efficiency by systematically filtering out less relevant patches while preserving fine-grained details. 

Our extensive experiments demonstrate that GFT achieves state-of-the-art performance across multiple benchmark datasets (FGVC, Aircraft, Food-101, and COCO) while maintaining efficiency. Compared to existing ViT-based approaches, our GFT provides a more interpretable, flexible and computationally efficient solution, effectively bridging the gap between global context understanding and localized feature extraction.

While GFT demonstrates strong performance, several limitations should be acknowledged.The patch selection mechanism, though efficient, may occasionally discard relevant information in highly similar classes where discriminative features are subtle or distributed. Additionally, the computational requirements, while improved over baseline models, still present challenges for deployment on resource-constrained edge devices.

Future work could explore scaling GFT to larger datasets and integrating multi-modal inputs (e.g., text or depth cues) to further enhance fine-grained recognition. Additionally, investigating real-time deployment optimizations could broaden its applicability in dynamic environments. Overall, GFT establishes a new benchmark in FGIC, offering a robust and adaptable framework for deployment of visual recognition systems in real-world scenarios.

\section{Code and Data Availability Statement}

All datasets are publicly cited and available. The code is open-sourced and can be found at: 
\url{https://github.com/sparcus-technologies/gradient-focus-transformer}

\section{Contributions}

\begin{itemize}
    \item \textbf{Boris Kriuk} introduced the idea, led the research, developed Methodology section, performed experiments, approved the final draft.
    
    \item \textbf{Shoaib Aslam} developed Related Works section, performed experiments, approved the final draft.
    
    \item \textbf{Fakhrutdinov Amir} developed Introduction and Experiments sections, performed experiments, approved the final draft.
    
    \item \textbf{Simranjit Kaur Gill} developed Introduction and Experiments sections, performed experiments, approved the final draft.
\end{itemize}

{\small
\bibliographystyle{ieeetr}
\bibliography{egbib}
}
\end{document}